\documentclass[runningheads]{llncs}
\usepackage[T1]{fontenc}
\usepackage{graphicx}
\usepackage{tabularx,ragged2e}
\usepackage{booktabs}
\usepackage{float}

\begin{document}
\title{Identifying Differential Equations to predict Blood Glucose using Sparse Identification of Nonlinear Systems}
\titlerunning{Prediction of Blood Glucose using SINDy}
%
%

\author{David Joedicke\inst{1}\orcidID{0000-0002-4970-2324} 
\and\\
Daniel Parra\inst{2}\orcidID{0000-0001-7460-7401} 
\and\\
Gabriel Kronberger\inst{1}\orcidID{0000-0002-3012-3189} 
\and\\
Stephan M. Winkler\inst{1}\orcidID{0000-0002-5196-4294} } 
%
%
\institute{Josef Ressel Centre for Symbolic Regression \\
Heuristic and Evolutionary Algorithms Laboratory\\
University of Applied Sciences Upper Austria, Hagenberg, Austria \email{david.joedicke@fh-ooe.at}
\and
Universidad Complutense de Madrid, Spain \\ \email{dparra@ucm.es}}
\maketitle              

\renewcommand{\thefootnote}{}
\footnotetext{\hspace{-0em}
Submitted manuscript to be published in \textit{Computer Aided Systems Theory - EUROCAST 2022: 18th International Conference, Las Palmas de Gran Canaria, Feb. 2022}.
}
\renewcommand\thefootnote{\arabic{footnote}}

\begin{abstract}
Describing dynamic medical systems using machine learning is a challenging topic with a wide range of applications. In this work, the possibility of modeling the blood glucose level of diabetic patients purely on the basis of measured data is described. A combination of the influencing variables insulin and calories are used to find an interpretable model. The absorption speed of external substances in the human body depends strongly on external influences, which is why time-shifts are added for the influencing variables. The focus is put on identifying the best time-shifts that provide robust models with good prediction accuracy that are independent of other unknown external influences. The modeling is based purely on the measured data using Sparse Identification of Nonlinear Dynamics. A differential equation is determined which, starting from an initial value, simulates blood glucose dynamics. By applying the best model to test data, we can show that it is possible to simulate the long-term blood glucose dynamics using differential equations and few, influencing variables.

\end{abstract}
\section{Background and Motivation}

Approximately 422 million people worldwide are diagnosed with diabetes and it accounts for an estimated 1.5 million deaths per year. Diabetes is a group of metabolic diseases with high blood sugar levels, which produces the symptoms of frequent urination, increased thirst, and increased hunger. Untreated, diabetes can cause many complications like ketoacidosis and nonketotic hyperosmolar coma. Serious long-term complications include heart disease, stroke, kidney failure, foot ulcers and damage to the eyes. Diabetes is due to either the pancreas not producing enough insulin, or the cells of the body not responding properly to the insulin produced. There are three main types of diabetes, which are different in its cause and symptoms. In this paper we use data of patients suffering from Type 1 diabetes, which results from the body's failure to produce enough insulin \cite{who}\cite{amdiabasso}.

\subsection{Dynamical systems and differential equations}
Numerous systems and processes can be represented by dynamic models. Whenever there is change over time, we speak of dynamic systems. Dynamic systems are often difficult to model, since the underlying dependencies between the parameters of a system are not known. Due to this discrepancy between diverse applications and complexity in the mapping, the modeling of dynamic systems is a research area with various approaches \cite{raol}. \\
In most applications, the differential equations of dynamic systems are not known. Instead, there are measurements of the individual variables. Our goal is to extract the differential equation system from the measurement data. Measurement errors, noise or unknown influences play a significant role here. In the case of modeling the blood glucose level, the course can be controlled by the supply of carbohydrates and insulin, unknown influences such as activity level, sleep quality or heart rate must be eliminated in the modeling step.

\section{Prior Work}
The evolution of blood glucose levels of diabetic patients has already been modelled using different machine learning methods such as support vector machines, gradient-boosting, or different types of neural networks. To better represent and analyse the dynamic behaviour of blood glucose, methods that provide less complex and more interpretable models are preferable \cite{vandoorn}. \\
Additionally to those static models the advantage of the usage of differential equations to predict the blood glucose dynamics is part of various research approaches. Gatewood et al. already introduced an idea to predict these dynamics using a fixed structure within a set of differential equations in 1970. They proposed to describe the change of the glucose value as $\frac{dg}{dt}=-m_1g-m_2h+J(t)$ and the net
hypoglycemic promoting hormone as $\frac{dh}{dt}=-m_2g+m_4h+K(t)$, where $m_1$ to $m_4$ are positive constants, $J$ and $K$ represent the entry rate of glucose and insulin and $t$ is time. They showed that it is possible to estimate the values for the $m$-parameters to describe the blood glucose level in the human body \cite{gatewood}.\\
Additionaly to this very simple approach multiple papers adapted the idea to improve the results. Chervoneva et al. proposed an approach to estimate the parameters of a mathematical model based on a small number of noisy observations using generalized smoothing \cite{chervoneva}. Shiang et al. performed those parameter estimation by minimizing the sum of squared residuals \cite{shiang}. The described approaches fit parameters of pre-defined structures to measured data, while this paper introduces a possibility to create accurate models without further specification of the structure of the differential equations. Those models are still interpretable since only few, important terms are used to describe the underlying dynamic. Additionally we particularly predict the long-term blood glucose dynamics for an entire day. The known approaches mainly focus on the short-term behaviour, predicting values up to two hours.\\

\section{Methodolgy}

\subsection{Data}
We use the \textit{OhioT1DM Dataset} \cite{ohiotdm1}, which includes the health and wellbeing over 8 weeks for 12 anonymized patients with type 1 diabetes. For each of the patients multiple health data and live events are measured. The following variables are used to predict the glucose value:

\begin{itemize}
    \item \textbf{Basal:} Long acting insulin, continuously infused
    \item \textbf{Bolus:} Short acting insulin, delivered to the patient at irregular time intervals
    \item \textbf{Meal:} Meals including the carbohydrate estimate for each meal
\end{itemize}

Combining all of the measured variables together creates a dataset with continuously measured data points every five minutes. There are two types of measurements: Influencing variables (basal, bolus and meals) can be influenced by the patients by taking insuline or eating something. Those impacts can be used to influence the blood glucose level directly. The glucose level itself can't be influenced directly, only indirect by the influencing variables. In this paper we describe the blood glucose level specifically for each patient, which is why we have used the specific dataset of a single test person, in which the data were recorded most completely. This dataset contains 17 days full of data. We are using 11 days for training and validation, the other 6 days are for testing the model.

\subsection{Feature Engineering}

\subsubsection{Absorption of the substances by the body:}
Since the dissolution of substances in the body occurs with a time delay and not abruptly, we preprocessed both bolus and carbohydrate values using the Berger function: \cite{berger} \\

\newenvironment{conditions}
  {\par\vspace{\abovedisplayskip}\noindent\begin{tabular}{>{$}l<{$} @{${}={}$} l}}
  {\end{tabular}\par\vspace{\belowdisplayskip}}

\begin{equation} \label{berger_fun}
   \frac{dA}{dt} = \frac{s \cdot t^s \cdot {T_{50}}^s \cdot D}{t \cdot ({T_{50}}^s+t^s)^2} - A
\end{equation}

\begin{equation} \label{berger_fun}
   T_{50} = a \cdot D \cdot b
\end{equation}

where $A$ is the plasma insulin, $D$ the insuline dose, $t$ the time after the injection, $s$ the time course of absorption and $a$ and $b$ are parameters to characterize dependency of $T_{50}$ on $D$. In our preprocessing we calculated the Berger function with $s=1.6$, $a=5.2$ and $b=41$

The Berger function converts a one-time recorded input such as the ingestion of insulin as well as carbs to such an extent that the uptake of the substance into the body does not take place abruptly but distributed over a longer period of time. The parameters from Equation (\ref{berger_fun}) determine exactly how the substance dissolves in the body.

\subsubsection{Time-Shifts} \label{time-shifts}
Additionally to the specified dissolution of the carbohydrates and the bolus in the body we implemented a time-shift to both the carbohydrates and the bolus value. Introducing these time-shifts allows us to dampen the impact of different activity levels across different days. For example, if a person exercises, the ingested carbohydrates dissolve more quickly in the body than if the person went a day without exercise. We implemented time-shifts for thirty minutes in the past, having a data-point every five minutes. This leads to six shifts for both the bolus and the carbohydrates. Our goal is to find an optimal time-shift for both the carbohydrates and the bolus value during the training phase and use those time-shifts for the test data.

\subsection{Algorithm}
Sparse Identification of Nonlinear Dynamical Systems (SINDy) uses sparsity techniques and machine learning to discover differential equations underlying a dynamical system. SINDy takes advantage of the fact, that most dynamical systems have only a few relevant terms, making them sparse in the space of possible functions. The numerical difference at each time step $t$ for each parameter is used to create a matrix $X$. For each parameter of $X$ a so called library $\Phi(X)$ with nonlinear functions of the column of $X$, such as constants, polynomials and trigonometric functions is created. Each column of $\Phi(X)$ represents a term for the right hand side of a differential equation. Using sparse regression only those terms with relevance to the system are kept \cite{brunton}.

\subsection{Grid Search}
To detect the best model several steps were performed:
\begin{itemize}
    \item Step 1: Find the best time-shift
    \item Step 2: Select best model with the detected time-shifts
    \item Step 3: Evaluate model with test data
\end{itemize}

The model for predicting the dynamics of glucose levels in blood should be as robust as possible to external influences, this is done as explained in Section \ref{time-shifts} by introducing time-shifts for the supply of carbohydrates as well as insulin. We define a model as robust that achieves the lowest error of the glucose value across all training days. We use the average absolute error per prediction as the error value for the measurement of the quality of a model.

\subsubsection{Step 1: Find most robust time-shift:}

In the first step, the best combination of the two defined time-shifts is to be found, i.e. for which shift of the two variables the most stable results are delivered. For this purpose, a model is created for each combination of carbohydrate shift and bolus shift for each of the training days. These models are then applied to all other combinations of time-shifts for all other training days. To find the best time-shift we used the following procedure: (1) Simulate blood glucose levels for all training days and time-shifts for each of the models. (2) Sort the results by the median error for each model and take the 10\% best results per time-shift. (3) Define the best time-shift as the one with the lowest mean absolute error across those 10\% of the best results.

\subsubsection{Step 2: Select best model:}
After determining which of the time-shifts are the most robust ones, one of the models from all the training days has to be selected as the final model. We choose the model which has the lowest average error among the other training days, including all of the time-shifts for this validation step. This leads to a model, which is robust against different time-shifts which might occur due to different behaviours of the body on different days. 

\subsubsection{Step 3: Evaluate model with test data:}
In the last step, the selected model is used to predict the blood glucose level of the days that were not used in the training. We took the first value of blood glucose level as the start value and predict the rest of the day integrating the differential equation selected in step 2. The quality of the predicted results is determined by the root mean squared error, the mean absolute error and the correlation. Each of the quality measurements has its advantages, using multiple of them allows a better comparison of the accuracy of the results. For a better interpretation of the results, we compare the determined glucose values with a simple, constant model (CM) in which we use the starting value as a constant value over the entire test range. 

\begin{figure}  
    \centering
    \includegraphics[width=13cm]{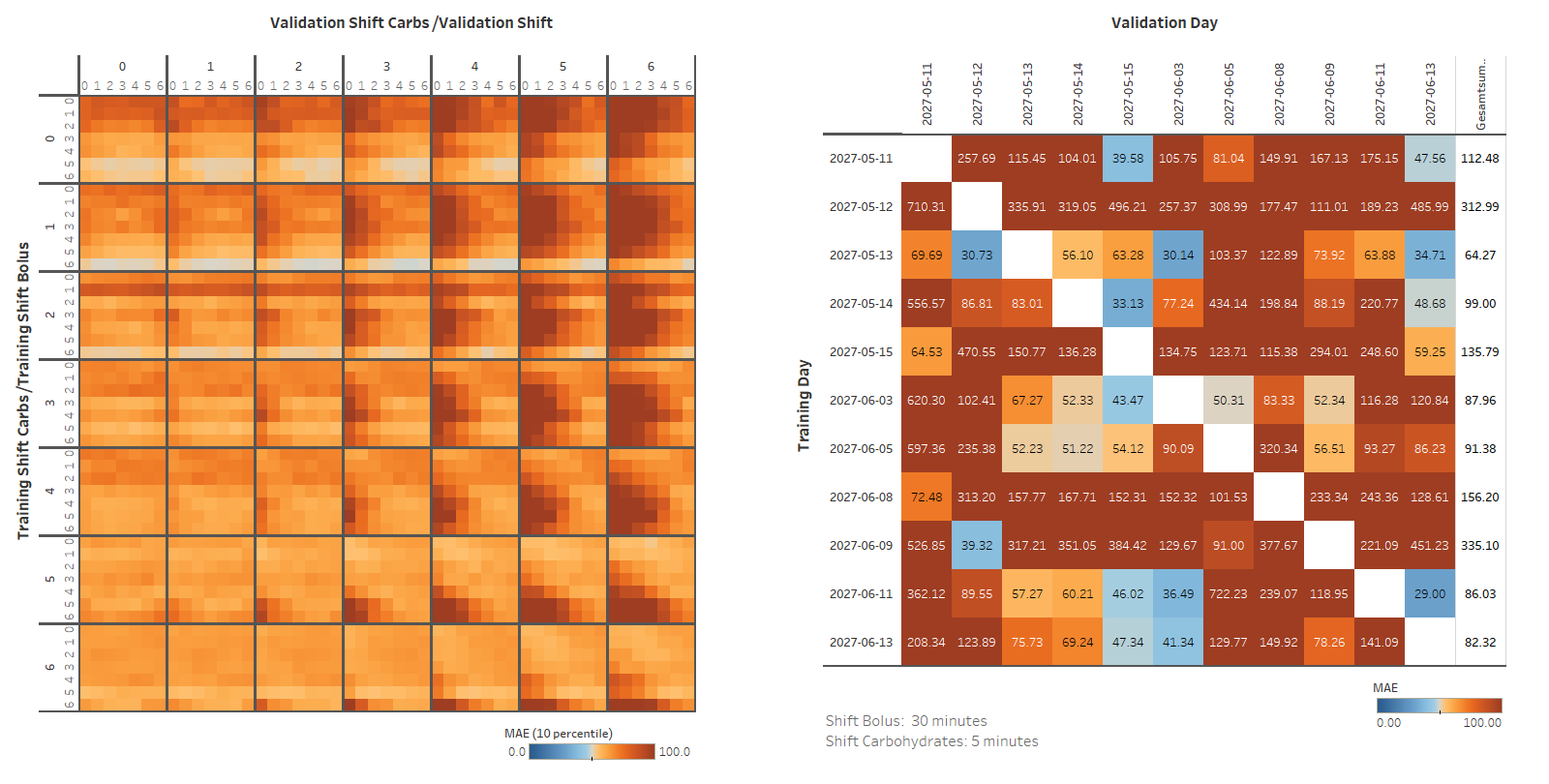}
    \caption{Quality of Models}
    \label{fig:model_error}
\end{figure}

\section{Results}
The result of each step is the input for the next step: The optimal time-shifts are used to select the best model, this differential equation is then applied to calculate the prediction for the test data.

\subsection{Best time-shift}

The left part of Figure \ref{fig:model_error} shows the mean absolute error of the 10\% best results for the combination of each time-shifts across all days. The best time-shifts evaluated across all training days are the following:
\begin{itemize}
    \item \textbf{Bolus:} 30 minutes
    \item \textbf{Carbohydrates:} 5 minutes
\end{itemize}

\subsection{Best Model}

Using the time-shifts from the previous step the best model is selected. The right part of Figure \ref{fig:model_error} shows the error of all the models for each days using the specified time-shifts. The best model is the model trained on the data of 13th of May 2027 which has the lowest validation error with an average mean absolute error of $64.27 \frac{mg}{dL}$. The differential equation of this model is:\\

\begin{equation}
   \frac{dG}{dT}=p_0 + p_1 \cdot b + p_2 \cdot C + p_3 \cdot G + p_4 \cdot B \cdot b + p_5 \cdot b^2 + p_6 \cdot b \cdot C + p_7 \cdot b \cdot G + p_8 \cdot C \cdot G + p_9 \cdot G^2
\end{equation}

\newcolumntype{R}[1]{>{\raggedleft\let\newline\\\arraybackslash\hspace{0pt}}m{#1}}

\begin{table}
\begin{tabular}{ m{2em} R{3.3em} R{6em} m{25em} } 

\textit{$p_0$} = & -1.14    & \textit{G} &= Glucose \\
\textit{$p_1$} = & 102.39   & \textit{C} &= Carbohydrates (bergerized, time-shift: 5 minutes) \\
\textit{$p_2$} = & -0.14    & \textit{B} &= Bolus (bergerized, time-shift: 30 minutes) \\
\textit{$p_3$} = & -7.69    & \textit{b} &= Basal\\
\textit{$p_4$} = & -0.21 \\
\textit{$p_5$} = & 648.80 \\
\textit{$p_6$} = & 12.80 \\
\textit{$p_7$} = & -185.42 \\
\textit{$p_8$} = & -0.96 \\
\textit{$p_9$} = & 11.39 \\
\end{tabular}
\end{table}

\subsection{Predicted Values}
\begin{figure}  
    \centering
    \includegraphics[width=10cm]{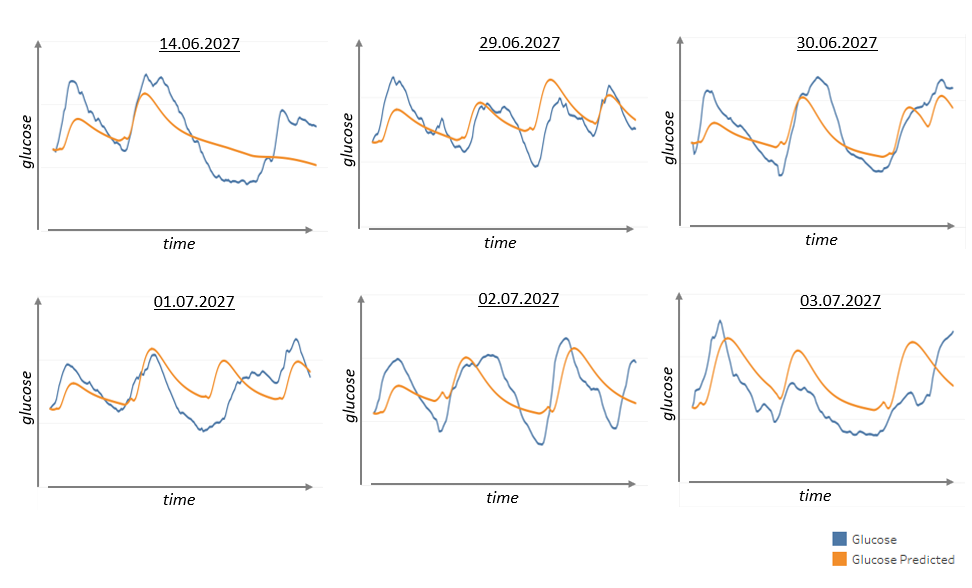}
    \caption{Exemplary results}
    \label{fig:exemplary_results}
\end{figure}

Equation (3) is used to predict the blood glucose level with only the start value of the blood glucose level and the influencing variables. Figure \ref{fig:exemplary_results} shows these results for all of the test days.

\subsection{Quality of results}

Table \ref{tab:overview results} shows the quality of the different test days compared to the baseline results of a constant model (CM). It is clear that by using SINDy, the prediction of blood glucose levels works much better than with a simple constant model.

\begin{table}[]
    \centering
    {\tabcolsep=0pt\def\arraystretch{1.5} \begin{tabularx}{300pt}{c *6{>{\Centering}X}}\toprule
    \multicolumn{1}{c}{} & \multicolumn{2}{c}{RMSE [mg/dL]} & \multicolumn{2}{c}{MAE [mg/dL]} & \multicolumn{1}{c}{R²}\\
    \cline{2-6}
    \multicolumn{1}{c}{Interval} & SINDy & CM & SINDy & CM & SINDy\\
    \hline
    14.06.2027 & 40.55 & 59.23 & 34.15 & 50.30 & 0.67\tabularnewline
    29.06.2027 & 28.41 & 46.30 & 20.92 & 28.41 & 0.50\tabularnewline
    30.06.2027 & 29.13 & 55.74 & 23.59 & 29.13 & 0.84\tabularnewline
    01.07.2027 & 35.01 & 48.42 & 28.16 & 40.59 & 0.39\tabularnewline
    02.07.2027 & 36.37 & 58.45 & 31.40 & 49.60 & 0.55\tabularnewline
    03.07.2027 & 49.17 & 50.41 & 45.68 & 37.89 & 0.56\tabularnewline 
    \hline
    Average & 36.44	& 53.09 & 30.65 & 39.32 & 0.59

    \end{tabularx}}
    \caption{Overview results}
    \label{tab:overview results}
\end{table}

\section{Summary and Outlook}
Modeling blood glucose levels with SINDy makes it possible to predict the long-term course of blood glucose levels in a diabetic patient. The model found is robust to possible external unknown influences and therefore allows to give a valid prediction about the future development of the blood glucose level. Although the overall prediction during an entire day is close to the actual values, the longer the prediction horizon is the higher the deviation is. Modeling of shorter time periods was not addressed in this paper; based on the good results over a longer observation horizon, the knowledge gained will be applied to other time intervals in future experiments. 
%
%
%
%

\end{document}